\documentclass[letterpaper]{article}

\usepackage{natbib} 
\usepackage{alifeconf}
\usepackage{url}
\usepackage{booktabs}
\usepackage{comment}
\usepackage{amsfonts}
\usepackage{amsmath}
\usepackage{hyperref}
\usepackage{cleveref}
\usepackage{float}

\makeatletter
\renewcommand\@fnsymbol[1]{}
\makeatother

\title{Risk-Aware Decision Policies for Agents Under Noisy Perception}

\author{
    David Szczecina\thanks{ \quad Submitted to the 2026 Conference on Artificial Life \mbox{(ALIFE2026)}} \\
    University of Waterloo, Canada \\
    david.szczecina@uwaterloo.ca
} 

\begin{document}

\maketitle

\begin{abstract}
    Perception in biological systems is inherently noisy, requiring organisms to make decisions under uncertainty where misclassification can be costly or fatal. 
    We present an Artificial Life predator-prey model of foraging under noisy perception, and compare agent performance when using various policies that take into account their noisy predictions.
    Through controlled experiments under both symmetric and asymmetric perceptual noise, we show that blindly trusting perceptual labels leads to catastrophic failure as noise increases, while uncertainty-aware strategies significantly improve survival and reduce fatal errors. We further observe qualitative regime shifts in behaviour, with agents transitioning from exploratory to conservative strategies as uncertainty increases. Our model links risk-sensitive foraging, ecological information use, and Artificial Life by showing that explicit information gathering can improve robustness when perception is unreliable. These results highlight the importance of uncertainty-aware decision-making and provide an interpretable artificial life analogue to robust learning with noisy labels.

\end{abstract}





\section{Introduction}

Living organisms rarely perceive their environments with certainty. Sensory information is noisy, partial, and context-dependent, shaped by distance, occlusion, camouflage, motion, and environmental variability, \cite{dall2005information, lima1990behavioral}. In predator-prey settings, perceptual errors are also strongly asymmetric, as mistaking food for danger may reduce energy intake, but mistaking danger for food can be fatal. Behavioural ecology and decision theory have long emphasized that such uncertainty does not merely degrade performance, but actively shapes action selection, information use, and risk sensitivity.

Artificial Life provides a natural framework for studying these dynamics through embodied agents interacting with structured environments. However, many foraging and predator-prey models assume near-perfect perception, abstracting away one of the defining constraints of real biological decision-making \cite{reynolds1987flocks, ray1991approach}. In this work, we model perception instead as a noisy probabilistic labelling process: agents do not observe the true identity of an entity directly, but receive uncertain beliefs about whether an entity is food or predator. This framing is biologically motivated, but it also mirrors a broader problem in artificial intelligence and machine learning. In noisy-label learning, models often treat observed labels as ground truth even when annotations are mislabelled, leading to harmful updates when prediction confidence conflicts with the given label \cite{song2022learning, northcutt2021pervasive, pleiss2020identifying}. Our simulation studies the behavioral analogue of this problem: when should an agent trust a noisy label, and when should it act skeptically? 

Rather than treating perception as direct access to ground-truth categories, we model it as an inference problem. Agents receive multiple noisy signals about an entity’s identity and must combine them to form a belief over whether the entity is food or a predator. This can be viewed as a simple form of Bayesian inference, where different sources of evidence contribute to a posterior belief that guides action selection \cite{knill2004bayesian}. We study this question in a two-dimensional foraging environment populated by agents, food resources, and predators. Agents maintain an internal energy state and must trade off energetic reward against survival risk under uncertain perception. We compare three decision strategies: a blind strategy that acts directly on current perceptual labels, a skeptical strategy that only commits to approach or avoidance when its inferred probability of food is sufficiently far from uncertainty, and otherwise abstains from action, and a verify strategy that can spend time and energy to gather additional information before committing to approach or avoidance. This allows us to examine not only whether perceptual noise reduces performance, but how different policies redistribute error across ecologically meaningful outcomes such as survival duration, predation and starvation.

Our results show that increasing perceptual noise produces clear behavioural regime shifts and exposes the weakness of blind trust in unreliable labels. Strategies that incorporate skepticism or verification better avoid catastrophic false approaches, although often at the cost of reduced foraging efficiency. By explicitly incorporating perceptual uncertainty into the decision-making process, this work contributes to core Artificial Life themes of robustness, adaptation, and emergence. More broadly, it offers a useful analogy for robust learning under label noise: when labels are unreliable and the costs of error are asymmetric, cautious decision rules can outperform blind optimization against corrupted signals.

\section{Related Works}

Research on foraging and predator–prey decision-making has consistently emphasized that organisms operate under uncertain, noisy, and often misleading sensory information, rather than direct access to ground-truth environmental states. Risk-sensitive foraging theory formalizes this by showing that behaviour depends not only on expected reward, but also on variance, internal state, and asymmetric costs of error \cite{stephens1986foraging, bateson1998risk, houston1999models}. In particular, ecological decision systems are shaped by the fact that false negatives (failing to detect a predator) are often far more costly than false positives (unnecessarily avoiding food), making uncertainty fundamentally asymmetric.

A large body of work in behavioural ecology has further demonstrated that prey must act under uncertain threat perception, often relying on heuristics when accurate classification is infeasible \cite{lima1990behavioral}. The predation risk allocation hypothesis extends this by showing that optimal behaviour depends on the temporal and structural properties of danger, implying that decision-making must adapt dynamically to uncertainty rather than assume fixed strategies \cite{lima1999temporal}. Importantly, these frameworks implicitly assume that perception itself is imperfect, but most formal and computational models do not treat perceptual noise as a primary variable.

The ecology of information provides a complementary perspective, arguing that behaviour is governed not only by environmental states but by how organisms acquire, interpret, and act on information of varying reliability \cite{dall2005information, schmidt2010ecology}. Information gathering is often costly in time, energy, or exposure, leading to trade-offs between acting immediately on uncertain signals and delaying action to improve certainty. This motivates modelling observation or verification as an explicit behavioural action, rather than assuming that perception is passively given.

Critically, many real-world predator–prey interactions are mediated by noisy, ambiguous, or deceptive signals. Predators frequently exploit sensory biases through aggressive mimicry, presenting cues that resemble food to lure prey, while prey species may employ defensive mimicry to resemble dangerous organisms \cite{dawkins1979arms, bates1862insect, ruxton2004avoidance}. These interactions highlight that perception is inherently probabilistic, and that misclassification is not rare noise but a structured ecological feature. Such asymmetries motivate modelling perception as a noisy labelling process, where observed categories may systematically deviate from true identity.

This issue becomes especially important in systems involving ambush or stationary predators, which are common across ecological domains (e.g., sit-and-wait predators such as spiders, snakes, or even plants like Venus flytraps) \cite{scharf2011functional}. Unlike actively chasing predators, ambush predators rely heavily on camouflage, immobility, and deceptive signalling, making them particularly difficult to detect and increasing the likelihood of false safety signals. In such environments, the perceptual challenge is not merely detecting motion or proximity, but correctly interpreting ambiguous cues that may resemble harmless or rewarding stimuli \cite{perry1997animal}. As a result, the cost asymmetry is amplified: failure to detect a stationary predator often results in immediate death, while overly cautious avoidance primarily incurs opportunity cost. This makes ambush-based systems a particularly relevant setting for studying risk-sensitive behaviour under perceptual uncertainty, and directly motivates the use of stationary predators in the present model.

Despite these ecological insights, many Artificial Life and agent-based predator–prey models assume that agents can reliably identify food and predators, or treat perception as a simplified proximity-based signal \cite{reynolds1987flocks, ray1991approach}. While this abstraction allows focus on spatial dynamics or collective behaviour, it omits a defining constraint of real biological systems: agents act on beliefs that may be incorrect. This leaves a research gap for the case where the observed category itself is uncertain, and where errors in perception directly drive survival outcomes.

The present work builds on past work by modelling perception as a noisy labelling process, the framework captures both the probabilistic nature of ecological signals and the asymmetric consequences of misclassification. This allows us to study a central question that spans behavioural ecology, Artificial Life, and decision theory: when does it become adaptive to question, delay, or verify uncertain perceptual information rather than act on it directly?

\section{Method} \label{sec:method}

\subsection{Simulation Environment}
We implement a two-dimensional agent-based environment with three entity types: agents, food, and predators. The environment is a bounded continuous space of fixed size, initialized each episode with fixed numbers of each entity type placed uniformly at random.

Each agent maintains a position $\mathbf{x}_i \in \mathbb{R}^2$ and an internal energy state $E_i$. At each timestep, agents incur a metabolic cost and may move toward or away from nearby entities. Energy increases when food is consumed and decreases through movement and time. Agents die either from predator contact (immediate death) or starvation when energy reaches zero.

At each timestep, agents detect nearby entities within a fixed observation radius and select a target based on proximity. The simulation proceeds for a fixed number of steps or until all agents have died.

\subsection{Agent State and Action Space}
Each agent is defined by its position $\mathbf{x}$, energy $E$, and perceptual estimates of nearby entities. At each timestep, the agent selects a target entity and executes one of three actions:

\textbf{Approach:} move toward the target with fixed step size, potentially consuming food if contact occurs.

\textbf{Avoid:} move away from the target, increasing distance from the perceived threat.

\textbf{Observe:} remain stationary and incur an energy cost $c_{\text{obs}}$, while improving the accuracy of the perceptual estimate.

If the agent abstains from action (e.g., under uncertainty), it performs a step in a random direction. Movement updates are deterministic given the chosen action, and interactions (food consumption or predator collision) occur when the agent enters a small radius around the target.

\subsection{Perceptual Model}
Agents do not directly observe the true class of an entity. Instead, each observation consists of two noisy information sources, a categorical label subject to misclassification, and a continuous feature signal.

We model perception as Bayesian inference over the latent class $y \in \{\text{food}, \text{predator}\}$. Given an observation consisting of a label $\ell$ and feature value $z$, the agent computes a posterior belief:
\[
P(y = \text{food} \mid \ell, z) \propto P(\ell \mid y) \, P(z \mid y) \, P(y).
\]
This formulation can be interpreted as integrating multiple noisy signals through additive contributions to the log-likelihood ratio.

\textbf{Label model.}
The observed label $\ell$ is a noisy version of the true class, with misclassification probability $\epsilon$:
\[
P(\ell = \text{food} \mid y = \text{food}) = 1 - \epsilon, 
\]
\[
P(\ell = \text{predator} \mid y = \text{food}) = \epsilon,
\]
and analogously for $y = \text{predator}$. In the asymmetric setting, these flip probabilities differ between classes.

\textbf{Feature model.}
The feature signal is drawn from a class-conditional Gaussian distribution:
\[
z \sim \mathcal{N}(\mu_y, \sigma^2),
\]
where $\mu_{\text{food}} = +\mu$ and $\mu_{\text{predator}} = -\mu$. The parameter $\mu$ controls the separation between classes, and $\sigma$ controls perceptual noise.

\textbf{Prior.}
We assume a uniform prior over classes, $P(y = \text{food}) = P(y = \text{predator}) = 0.5$.

The posterior is then normalized to obtain a scalar belief:
\[
p_{\text{food}} = P(y = \text{food} \mid \ell, z).
\]

\textbf{Observation action.}
The observe action generates a new observation $(\ell', z')$ with reduced label noise and lower feature variance, corresponding to improved likelihood reliability. The agent recomputes its posterior belief using the updated observation.

\subsection{Behavioural Policies}
We evaluate three fixed policies that differ in how they respond to perceptual uncertainty.

\textbf{Blind policy.}
The agent selects an action based on the maximum probability label:
\[
\text{action} =
\begin{cases}
\text{approach} & \text{if } p_{\text{food}} > p_{\text{pred}} \\
\text{avoid} & \text{otherwise}
\end{cases}
\]
This policy assumes perceptual labels are reliable and does not account for uncertainty.

\textbf{Skeptical policy.}
The agent introduces a confidence threshold $\tau$.
The agent computes a posterior probability $p_{\text{food}}$ and applies a confidence threshold $\tau$ around uncertainty. Actions are taken only when the predicted class exceeds this threshold:

\[\text{approach if } p_{\text{food}} > \tau, \quad \text{avoid if } p_{\text{pred}} > \tau\]

Otherwise, the agent defaults to avoidance or inaction. This biases behaviour toward safety under uncertainty.

\textbf{Verify policy.}
The agent performs an observation step when uncertainty is high:

\[\text{observe if } \max(p_{\text{food}}, p_{\text{pred}}) < \tau\]

Observation incurs a higher energy cost but produces a more reliable observation, after which the agent recomputes its posterior belief and selects an action using the same decision rule as the skeptical policy.

\section{Experiments}

\paragraph{Experimental Setup}
We evaluate agent performance under varying levels of perceptual noise in a two-dimensional environment. Each episode is initialized with 50 agents, 150 food items, and 20 predators placed uniformly at random in a bounded $100 \times 100$ space. Episodes run for a maximum of 750 time steps or until all agents have died.

Agents incur an energy cost for movement and time, gain energy by consuming food, and die either from predator contact or starvation. At each time step, agents select the nearest entity within a fixed observation radius and act according to their policy.

Perceptual noise is controlled by a parameter $\epsilon \in [0, 0.5]$, which determines the probability of label misclassification. In the symmetric setting, both classes are flipped with probability $\epsilon$. In the asymmetric setting, predator-to-food errors occur with higher probability, such that $P(\text{pred} \rightarrow \text{food}) = 1.5 \epsilon$, simulating a more hostile environment where predators better mimic the appearance of food sources to lure prey.

\paragraph{Policies Compared}
We compare three fixed behavioural policies: blind, skeptical, and verify.

The blind policy acts directly on the observed label without accounting for uncertainty. The skeptical policy thresholds the posterior belief $p_{\text{food}}$, only committing to actions when confidence exceeds fixed margins around uncertainty, $\tau = 0.65$, representing a minimum 65\% confidence. The verify policy extends this by performing an observation step when uncertainty is high, incurring a cost in time and energy to obtain more reliable perceptual evidence before acting.

All policies share the same underlying perceptual model and differ only in how they map beliefs to actions.

\paragraph{Evaluation Protocol}
We evaluate performance using the average number of time steps an agent remains alive, and the average number of food items consumed per agent, both of which agents try to maximize. 
Additionally we consider fraction of agent deaths due to energy depletion (starvation), the fraction of agent deaths caused by predators (predation), as well as the frequency with which agents approach predators due to incorrect perception (false approach rate).
These metrics capture both overall performance and the mechanisms underlying failure under uncertainty.

For each policy and noise level, we run simulations across 20 independent seeds and report the mean of each metric and with standard deviation shown as ribbon in all figures. Performance is analyzed as a function of noise magnitude to assess how increasing perceptual uncertainty affects behaviour and to compare robustness across policies.

We vary only the level of perceptual noise while keeping all other parameters fixed, allowing us to isolate the effect of uncertainty on agent behaviour.

\section{Results}

All figures feature averaged values over 20 seeds and shaded regions indicating $\pm 1$ standard deviation.

\subsection{Performance under Increasing Noise}

\begin{figure}[H]
    \centering
    \includegraphics[width=\linewidth]{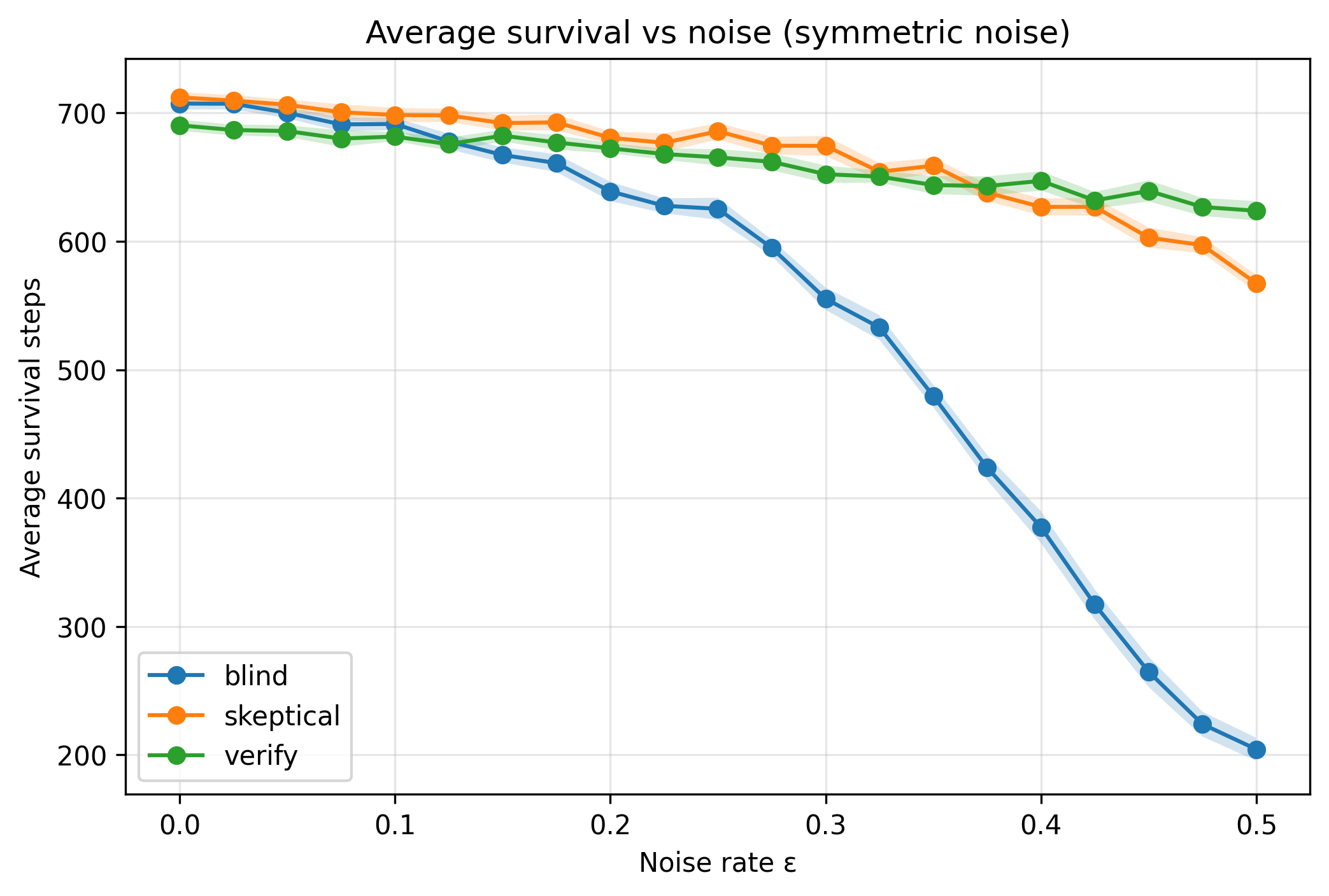}

    \vspace{-2mm}
    
    \caption{Average survival time (in time steps) as a function of perceptual noise $\epsilon$ under symmetric conditions. At low noise levels, all policies perform similarly. As noise increases, the blind policy collapses rapidly, while skeptical and verify policies degrade more gradually. The verify policy consistently achieves the highest survival, demonstrating improved robustness to noisy perception.
}
    \label{fig:survival_sym}
\end{figure}

Figure~\ref{fig:survival_sym} shows average survival time as a function of perceptual noise under symmetric conditions. At low noise levels ($\epsilon \le 0.1$), all policies achieve similar survival ($\sim700$ steps), indicating that when perceptual signals are reliable, even simple decision rules are sufficient.

As noise increases, a clear divergence emerges. The blind policy exhibits a sharp decline in survival, from $\sim 650$ at $\epsilon = 0.2$ to just 200 steps by $\epsilon = 0.5$. In contrast, both the skeptical and verify policies degrade more gradually, with the verify policy maintaining the highest survival across all noise levels. This demonstrates that policies which account for uncertainty are significantly more robust to perceptual corruption.

\begin{figure}[H]
    \centering
    \includegraphics[width=\linewidth]{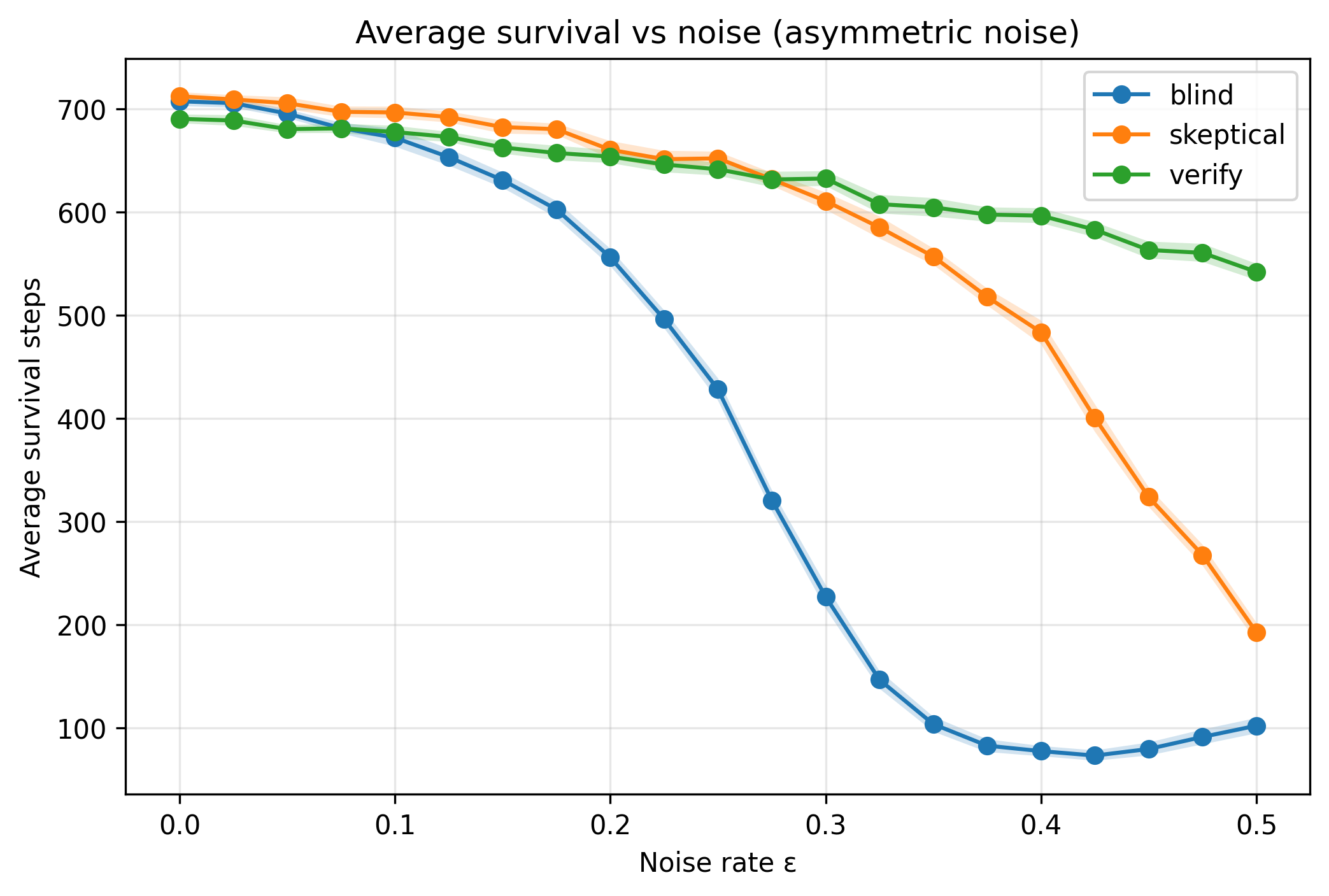}

    \vspace{-2mm}
    
    \caption{Average survival time (in time steps) as a function of perceptual noise $\epsilon$ under asymmetric conditions. At low noise levels, all policies perform similarly. As noise increases, the blind policy collapses rapidly, while skeptical and verify policies degrade more gradually. The verify policy consistently achieves the highest survival, demonstrating improved robustness to noisy perception.
}
    \label{fig:survival_asym}
\end{figure}

Under asymmetric noise where $P(\text{pred} \rightarrow \text{food}) = 1.5 \epsilon$, similar trends are observed, though the degradation is more sudden as seen in \Cref{fig:survival_asym}. The blind policy still performs worst, but the gap between policies is larger compared to the symmetric case. Additionally the skeptical policy begins to significantly degrades at higher noise rates, resulting in comparable results to the blind policy at $\epsilon \approx 0.5$. This indicates that asymmetric noise amplifies the consequences of misclassification, making robustness strategies more critical when one type of error (e.g., predator misclassification) is more costly.

\begin{figure}[h]
    \centering
    \includegraphics[width=\linewidth]{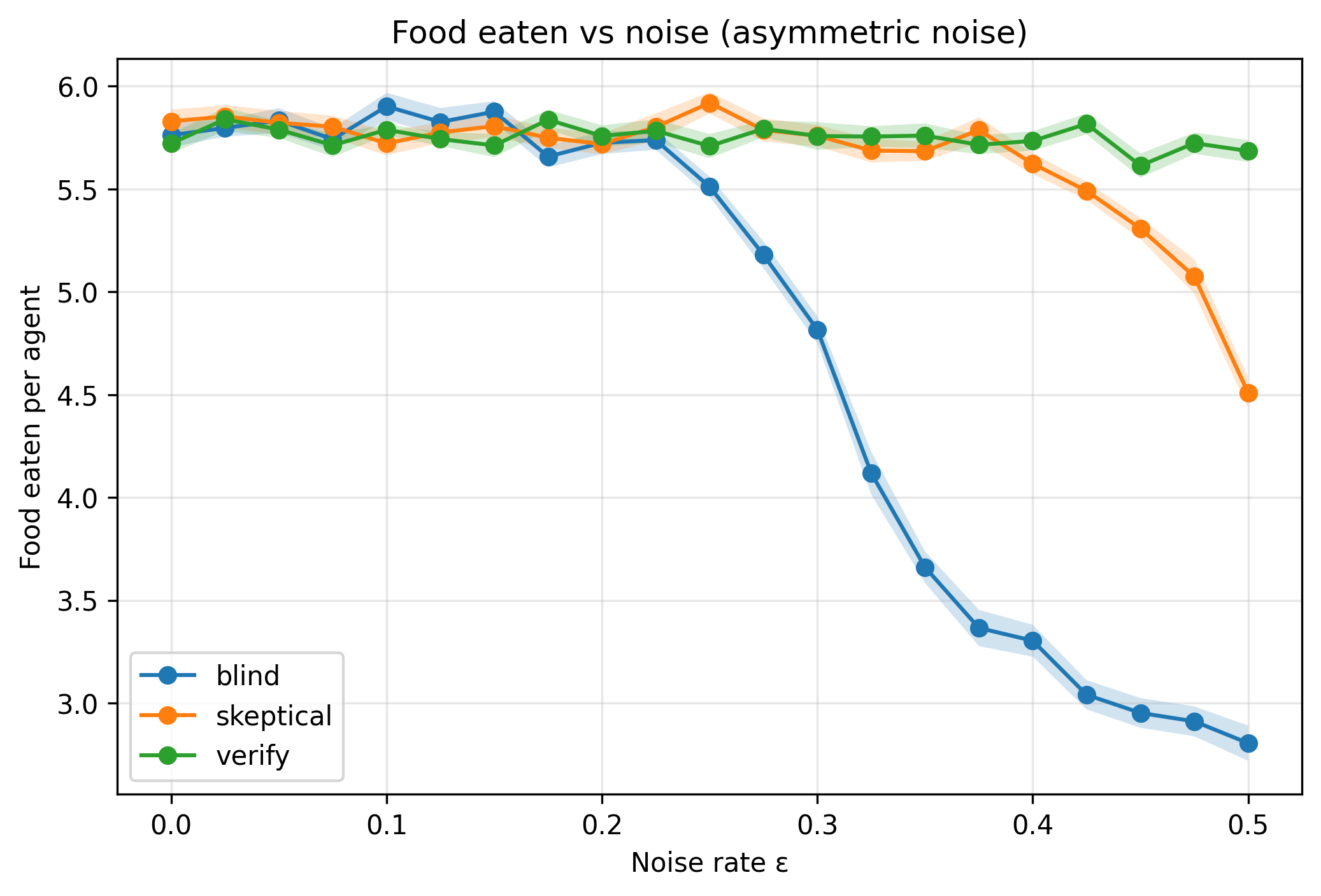}
    \caption{Average number of food items consumed per agent as a function of perceptual noise $\epsilon$ under asymmetric conditions. All policies achieve similar food intake at low noise levels. As noise increases, the blind policy initially maintains high food consumption but collapses due to increased predation, while skeptical policy exhibit a gradual decline, and the verify policy remains robust to the noisy perception. This illustrates the trade-off between resource acquisition and robustness to uncertainty.
}
    \label{fig:food_mean_asym}
\end{figure}

Graphing the average amount of food eaten per agent in \Cref{fig:food_mean_asym}, we see similar trends as the average survival duration. As noise rates increase, There is a sharp decline in performance of agents operating with the blind decision policy after around $\epsilon \approx 0.2$, and the performance of agents using the skeptical policy noticeably worsens at higher noise rates in the asymmetric noise experiments. In both situations the verify policy is significantly more robust to the noisy perception, ensuring agents live longer and consume more food on average.

\subsection{Failure Modes: Predation vs. Starvation}

\begin{figure}[h]
    \centering
    \includegraphics[width=\linewidth]{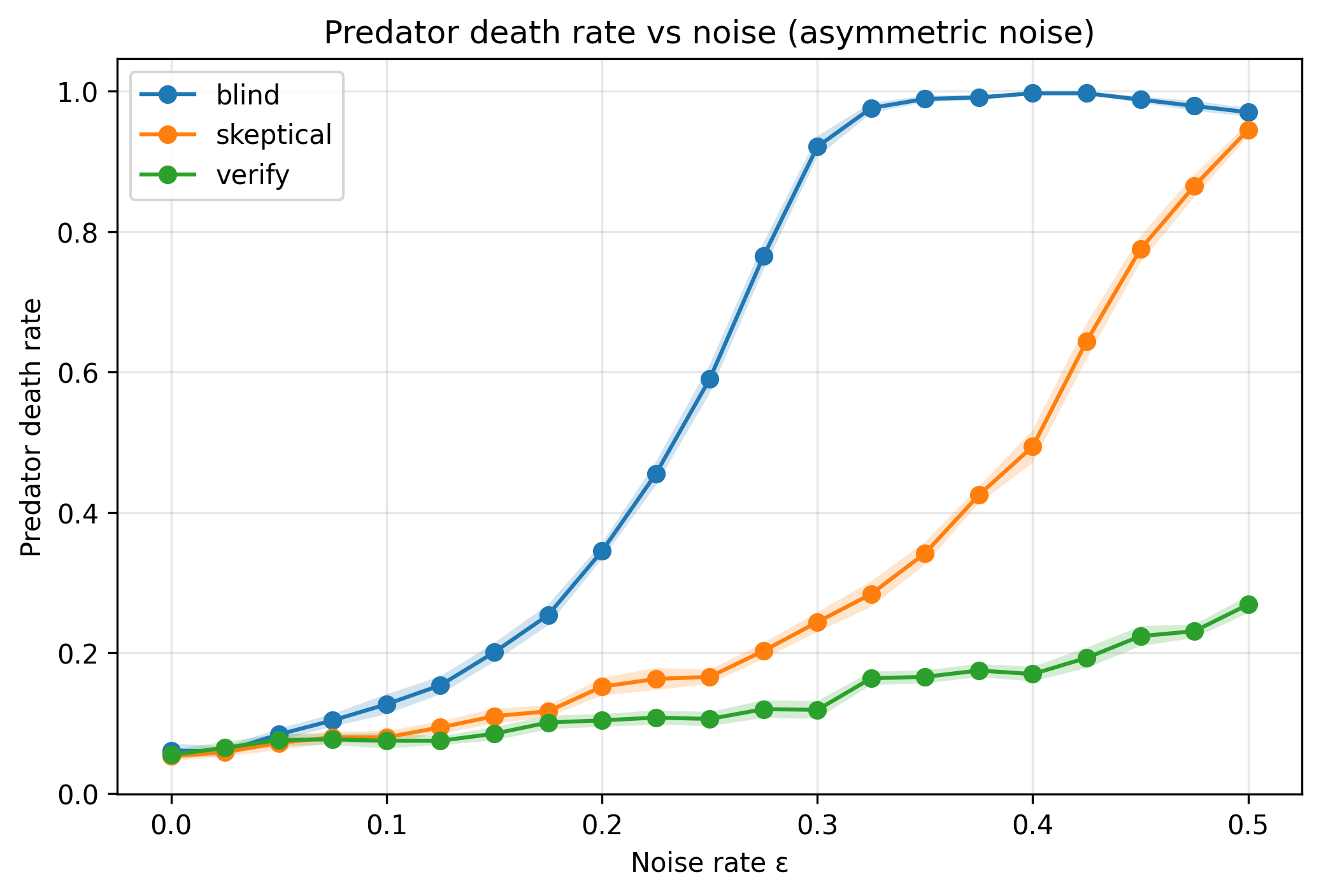}
    \caption{Fraction of agent deaths due to predation as a function of perceptual noise $\epsilon$ under asymmetric conditions. The blind policy shows a sharp increase in predator-induced deaths, approaching near-total failure at high noise levels. In contrast, skeptical and verify policies substantially reduce predation, with the verify policy maintaining the lowest death rates across all noise levels.
}
    \label{fig:pred_death_asym}
\end{figure}

To understand the source of performance degradation, Figure~\ref{fig:pred_death_asym} shows the fraction of deaths due to predation. The blind policy exhibits a dramatic increase in predator deaths as noise increases, approaching near-total failure at high noise levels. This indicates that agents frequently misclassify predators as food and approach them.

In contrast, the skeptical and verify policies substantially reduce predator deaths. The verify policy in particular maintains low predation rates even at high noise, suggesting that explicit information gathering effectively mitigates catastrophic errors.

\begin{figure}[H]
    \centering
    \includegraphics[width=\linewidth]{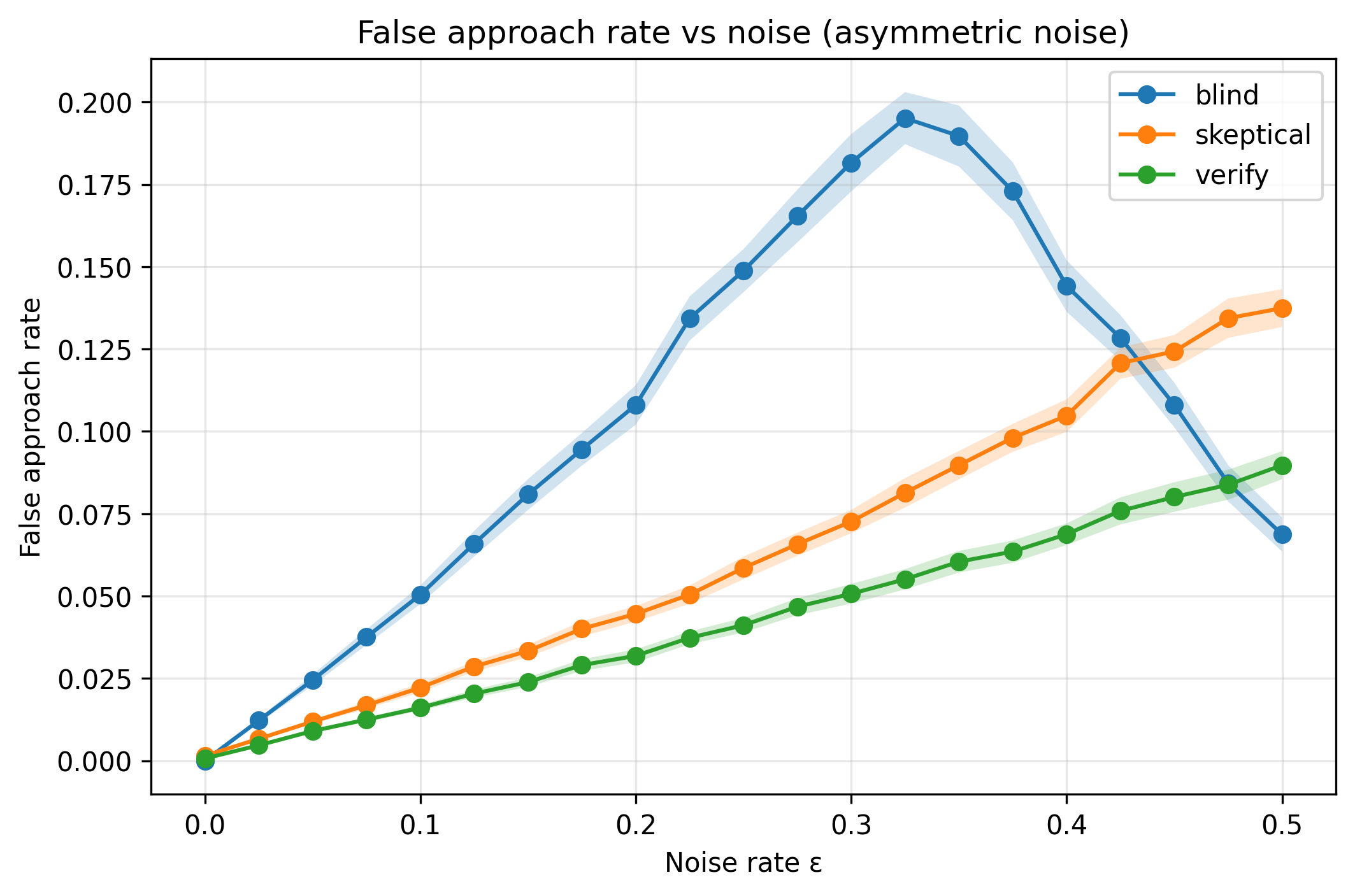}
    \caption{Rate of false approach actions (approaching predators due to misclassification) as a function of perceptual noise $\epsilon$. The blind policy exhibits a strong increase in false approach rate with noise, indicating systematic decision errors. Skeptical and verify policies significantly reduce these errors, with the verify policy achieving the lowest false approach rate.
}
    \label{fig:false_approach_asym}
\end{figure}

\subsection{False Approach Errors}

Figure~\ref{fig:false_approach_asym} illustrates the rate of false approach actions, where agents move toward predators due to incorrect perception. This metric directly captures the mechanism underlying performance collapse.

The blind policy exhibits a steep increase in false approach rate with noise, peaking at intermediate noise levels before declining as agents die more quickly. In contrast, the skeptical policy reduces this error rate, while the verify policy achieves the lowest false approach rate across all noise levels.
This result highlights the key failure mode of blind decision-making: when labels are noisy, acting directly on them leads to systematic and catastrophic errors. Policies that incorporate uncertainty either by thresholding (skeptical) or active verification can significantly reduce these errors.

\subsection{Trade-off Between Safety and Resource Acquisition}

\begin{figure}[h]
    \centering
    \includegraphics[width=\linewidth]{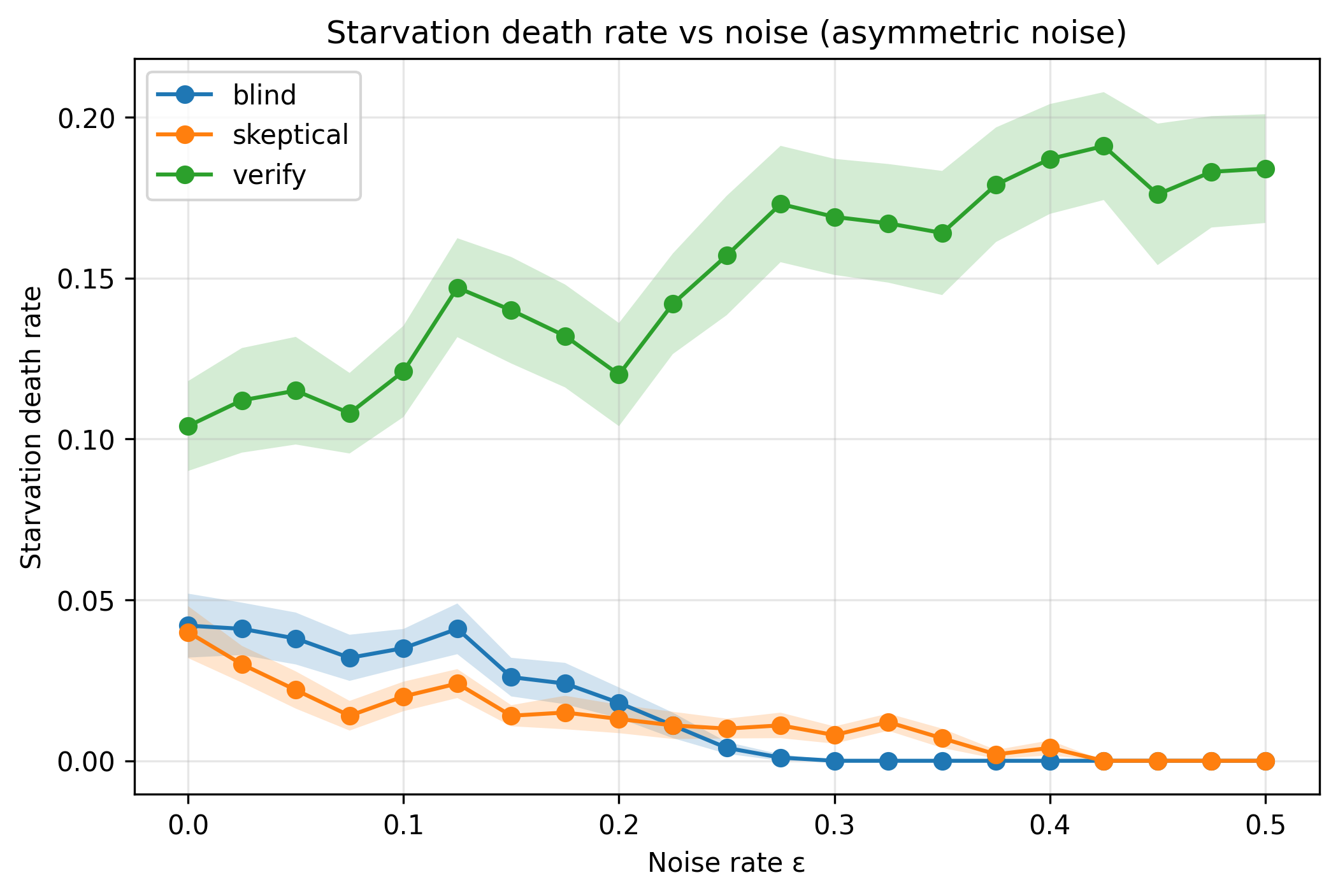}
    \caption{Fraction of agent deaths due to starvation as a function of perceptual noise $\epsilon$ under asymmetric conditions. The verify policy exhibits higher starvation rates as noise increases, reflecting more conservative behaviour and missed feeding opportunities. This highlights the trade-off between avoiding predation and acquiring resources.
}
    \label{fig:starve_death_rate_asym}
\end{figure}

While uncertainty-aware policies reduce catastrophic errors, they introduce a trade-off between preventing predation and resource acquisition. As shown in \Cref{fig:starve_death_rate_asym} results, the verify policy experiences significantly higher starvation rates as the noise rate increases. By choosing to spend additional time trying to verify whether an entity is food or a predator, it sacrifices energy and time reflecting missed opportunities for food acquisition. With higher noise rates agents choose to verify more often resulting in more starvation deaths, however, this trade-off is beneficial overall, as avoiding lethal mistakes leads to longer survival and on average more food consumed.

\section{Discussion}

\paragraph{Uncertainty and Risk}

The results demonstrate that perceptual uncertainty does not simply degrade performance, but fundamentally alters the structure of decision-making. At low noise levels, all policies perform similarly, indicating that when perceptual signals are reliable, simple reactive strategies are sufficient. However, as noise increases, performance becomes dominated not by average reward, but by the cost of rare but catastrophic errors.

The primary failure mechanism is the increase in false approach errors, where agents misclassify predators as food and act on this incorrect belief. This induces a clear causal chain: increasing noise degrades posterior belief accuracy, the blind policy commits to actions without accounting for uncertainty, and asymmetric error costs amplify the consequences of these mistakes. As a result, even a moderate increase in misclassification probability leads to a sharp collapse in survival. This highlights that under uncertainty, decision quality is governed less by typical outcomes and more by the tail risk of high-cost errors.

Importantly, the results suggest the presence of a behavioural regime shift. At low noise, agents operate in a reward-driven regime where maximizing food intake is sufficient. Beyond a critical noise threshold (approximately $\epsilon \approx 0.2–0.3$), behaviour transitions to a risk-dominated regime, where survival depends primarily on avoiding catastrophic mistakes. In this regime, policies that fail to account for uncertainty rapidly become non-viable.

\paragraph{Trade-offs and Redistribution of Failure Modes}
While uncertainty-aware policies improve robustness, they do not eliminate failure, but instead redistribute it. The blind policy minimizes conservatism and maintains high food acquisition at low noise, but suffers from rapidly increasing predation as noise increases. In contrast, the skeptical and verify policies reduce exposure to predators, but at the cost of reduced foraging efficiency.

This trade-off is most clearly observed in the verify policy, which exhibits increased starvation rates at higher noise levels. By investing time and energy into improving perceptual certainty, agents reduce the likelihood of catastrophic predation, but incur opportunity costs that lead to missed feeding events. The skeptical policy occupies an intermediate position, reducing high-risk actions without incurring the full cost of active information gathering.

These results highlight a key principle: robustness emerges not from eliminating errors, but from shifting the distribution of failure modes. Under asymmetric costs, it is advantageous to accept more frequent low-cost failures (e.g., starvation) in order to avoid rare but high-cost outcomes (e.g., predation). This reframes robustness as a problem of risk management rather than performance maximization.

\paragraph{Implications for Noisy-Label Learning}
The observed dynamics closely parallel supervised learning under label noise. In standard training pipelines, models often treat labels as ground truth even when they are incorrect, leading to harmful updates when predictions confidently disagree with supervision. This is directly analogous to the blind policy, which acts on noisy perceptual labels without accounting for uncertainty.

In this analogy, false approach errors correspond to fitting mislabeled samples: the model updates strongly in the wrong direction, despite high-confidence disagreement. As in the simulation, these errors are not merely random noise, but systematic and high-impact, and can dominate overall performance as noise increases.
The skeptical and verify policies provide behavioural analogues of robust learning strategies. Skepticism corresponds to down-weighting or ignoring uncertain or conflicting samples, similar to confidence-based filtering or robust loss functions \cite{pellegrino2024loss}. Verification corresponds to re-evaluating or refining labels, analogous to relabeling, semi-supervised learning, or incorporating additional signals before updating the model.

More broadly, these results suggest that the core issue in noisy-label learning is not noise itself, but the interaction between noise and overconfident decision-making. When systems commit strongly to potentially corrupted supervision, errors propagate and accumulate \cite{song2022learning, northcutt2021pervasive, zhang2017understanding}. Treating labels as uncertain, either by reducing their influence or by actively refining them, leads to significantly more robust outcomes.

\paragraph{Role of Information Gathering}
The verify policy highlights the importance of treating information acquisition as an explicit and adaptive component of decision-making. Rather than passively accepting noisy observations, agents can choose to invest resources to improve belief accuracy before acting. The increased frequency of observation at higher noise levels indicates that the value of information is strongly context-dependent.
Information gathering becomes most beneficial when uncertainty is high and the cost of incorrect action is severe. 

This aligns with theoretical perspectives in ecology and decision theory, where agents balance the expected value of information against its cost. In the present setting, observation effectively reduces uncertainty in the posterior belief, allowing agents to avoid catastrophic errors even in highly noisy environments.
These findings suggest that robust systems should not only incorporate uncertainty into their decision rules, but also dynamically control when to seek additional information. Fixed strategies may be insufficient in environments where uncertainty varies over time or across contexts.

\paragraph{Policy Optimality Across Noise Regimes}
No single policy is uniformly optimal across all noise levels. Instead, the results suggest that different strategies are appropriate under different uncertainty regimes. At low noise, the blind policy performs well due to its simplicity and lack of overhead. At moderate noise levels, the skeptical policy provides a strong balance between efficiency and safety by avoiding uncertain decisions without incurring additional costs. At high noise levels, the verify policy becomes dominant, as active information gathering is necessary to prevent catastrophic errors.
This highlights that robustness is inherently context-dependent. Effective strategies must adapt not only to the level of uncertainty, but also to the relative costs of different types of errors. Systems that fail to adapt their behaviour across regimes may perform well under specific conditions but degrade rapidly when those conditions change.

\paragraph{Limitations and Future Directions}

This study isolates the effects of perceptual noise using a simplified environment and fixed behavioural policies. Agents do not learn or adapt over time, and perception is modelled using a relatively simple probabilistic framework. As a result, the policies evaluated here represent hand-designed strategies rather than emergent behaviours.
Future work could extend this framework by incorporating learning-based agents, allowing strategies such as skepticism and verification to emerge through optimization. More complex environments, including spatial structure, evolving predators, or richer perceptual signals, would further test the generality of these findings. Additionally, extending the analogy to machine learning systems by explicitly integrating robust training methods or adaptive label correction mechanisms would provide a more direct connection between behavioural and algorithmic robustness.
\section{Conclusion}

This work investigated decision-making under perceptual uncertainty through a risk-sensitive foraging simulation. We showed that while simple, reactive strategies perform well in low-noise environments, their performance collapses as noise increases due to the accumulation of high-cost errors. In particular, agents that blindly trust noisy signals suffer from catastrophic failures, driven by rare but severe misclassifications.

In contrast, policies that incorporate uncertainty, either by acting conservatively or by actively improving their beliefs, demonstrate significantly greater robustness. These strategies do not eliminate errors, but instead shift behaviour toward avoiding high-risk outcomes, even at the cost of reduced efficiency. This highlights that under uncertainty, effective decision-making is governed not by maximizing reward, but by managing risk.

Beyond the simulation, these findings provide a conceptual parallel to supervised learning under label noise. Standard training procedures often treat labels as ground truth, leading to harmful updates when labels are incorrect. Our results suggest that incorporating skepticism toward supervision, through reduced reliance on uncertain signals or mechanisms for re-evaluation, can improve robustness in noisy settings.

Overall, this work emphasizes that robustness emerges from explicitly accounting for uncertainty. Systems that blindly trust their inputs may perform well under ideal conditions, but degrade rapidly when those inputs become unreliable. Designing agents and learning algorithms that recognize and adapt to uncertainty is therefore essential for reliable performance in real-world environments.



\footnotesize
\bibliographystyle{apalike}
\bibliography{references.bib}

\end{document}